# NPCS AS PEOPLE, TOO: THE EXTREME AI PERSONALITY ENGINE

Jeffrey Georgeson
Quantum Tiger Games, LLC
3431 W 17th Ave
Denver, CO 80204 USA
E-mail: jg@quantumtigergames.com

Christopher Child
City University London
Northampton Square
London EC1V 0HB, UK
E-mail: c.child@city.ac.uk

**KEYWORDS**

artificial intelligence, personality, non-player characters, video games

**ABSTRACT**

PK Dick once asked "Do Androids Dream of Electric Sheep?" In video games, a similar question could be asked of non-player characters: Do NPCs have dreams? Can they live and change as humans do? Can NPCs have personalities, and can these develop through interactions with players, other NPCs, and the world around them? Despite advances in personality AI for games, most NPCs are still undeveloped and undeveloping, reacting with flat affect and predictable routines that make them far less than human—in fact, they become little more than bits of the scenery that give out parcels of information. This need not be the case. Extreme AI, a psychology-based personality engine, creates adaptive NPC personalities. Originally developed as part of the thesis "NPCs as People: Using Databases and Behaviour Trees to Give Non-Player Characters Personality," Extreme AI is now a fully functioning personality engine using all thirty facets of the Five Factor model of personality and an AI system that is live throughout gameplay. This paper discusses the research leading to Extreme AI; develops the ideas found in that thesis; discusses the development of other personality engines; and provides examples of Extreme AI's use in two game demos.

## INTRODUCTION: THEN AND NOW

Non-player characters include all the inhabitants of the game world who aren't being played by human beings; they are, in effect, the original virtual actors. Their roles include everything from the bit parts (man in crowd, woman in shop) to main characters (player's confidante, love interest, main villain), yet they rarely have any autonomy or development and do not react to player actions or world events as real people would. At best, they change as the script dictates. At worst, they merely repeat the same tidbit of information again and again, no matter that the world around them is going down in flames.

The original 2011 project (unnamed at the time; referred to in this paper as the prototype personality engine, or PPE) sought to change this, allowing NPCs to be "utilised much more effectively and realistically—made, in fact, more human by giving them personalities that change over time and through interactions with the players, other NPCs, and the game world," giving them "added realism [that] would add depth and flavour where before there was only cardboard two-dimensionality" (Georgeson 2011). The objectives of the project included (as stated in the original):

- creating support NPCs whose personalities develop realistically over time depending on interactions with a player in a game environment
- comparing these NPCs to a set of unchanging NPCs whose reactions are controlled by random chance

As the PPE was developed into Extreme AI (ExAI), the system was further refined and developed. Its primary objectives became:

- creating more potential interactions for developers to call (originally there were but three—kindness, annoyance, and intimidation—which worked well for the PPE but were far too limiting for realism)
- speeding up the system so that complex interactions didn't noticeably slow down action in a game
- moving away from using a database for storage so that web- and mobile-based games could utilize the system

This paper summarizes the research and original work on the PPE, personality engine development since that time, and the current capabilities of ExAI.

## PERSONALITY ENGINES AND ACADEME

While limited in scope, the PPE was, at the time, the only one of its kind. Given increases in computational power, AI could be (and sometimes is) provided with increased resources, but it is still given short shrift compared to graphics (Doherty & O'Riordan 2006; Lemaitre, Lourdeaux, & Chopinaud Jan 2015). As stated in 2011:

> For example, *Final Fantasy XII* (Square Enix 2006) NPCs look very realistic, but they react in ways similar to such characters in the first Final Fantasy games in the 1990s, with possibly only a very basic finite state machine guiding the characters, and this is the same in role-playing games generally: When the player approaches the NPC for the first time, the character says something about the red dragons guarding the Great Treasure—and repeats the same thing every time the player comes near. After the player has completed this quest and returns to tell the NPC about it, the NPC may (or may not) be complex enough to say something about being happy no longer having to deal with the dragons.

Usually even this sort of thing does not iterate further; this NPC is done imparting information, and now will blithely blither on about the joys of a world without red dragons whenever prodded. There is no personality at all.

While scripting languages can provide more complex-seeming interactions, ultimately such step-wise scripting does not "grow" characters in terms of changing personalities or allow for natural changes in interactions, as the NPC never "feels" why she is reacting in a certain way; she is instead but a robot, following a preset directive along an if-then pathway. Such scripting "tend[s] to constrain [NPCs] to a set of fixed behaviours which they cannot evolve in time with the world in which they dwell" (Merrick & Maher 2006), and these behaviors are "hard to extend, maintain and learn" (Le Hy et al. 2004). (More recent research agrees; see Sales, Clua, de Oliveira, & Paes 2013.)

Add to this that many companies still believe that the AI could exhibit "inferior behavior" (Spronck et al. 2005)—possibly going mad, not revealing vital information when asked, etc.—and there are many motivations against developing more realistic personality AI, especially pre-2012 (and still in large part today).

Still, even prior to 2011, there were some projects attempting to create some form of personality system. Mac Namee (2004) creates an architecture to "drive the behaviours of *non-player support characters* in *character-centric* computer games" (italics in original). However, he uses Eysenck's outdated "two-dimensional classification" of personality (from 1965); this is coupled with a "mood model" (Lang 1995) measuring mood across two axes, valence and arousal, and also with a "relationship model" adding a "Level of Interest" value "indicating how interested one character is in another" (Mac Namee 2004). While a workable system, it has several drawbacks, including the use of a limited number of personality traits (extroversion/introversion and neuroticism/stability) that do not change once set. Also, while feelings toward specific characters can change, there is no allowance for these feelings to change the NPC's personality or feelings in general (e.g., feeling an initial wariness toward all strangers because several interactions have already gone awry) or for changes caused by non-characters (e.g., the character's village burns to the ground). Tied into the unchanging traits is the use of trained artificial neural networks (ANNs), which are trained in development and are turned off by the time players see the game—an efficient use of ANNs, but again leading to non-adaptive AI.

A model closer to ExAI's and very similar to that used by Pixel Crushers' recent Love/Hate engine (2015) is Li & MacDonnell's (2008) use of the Five Factor Model (Goldberg 1993; John et al. 2010; McCrae & Costa 2010; and DeYoung 2010), which posits that human personality consists of five factors: Openness, Conscientiousness, Extraversion, Agreeableness, and Neuroticism (OCEAN). These factors are further divided into 30 underlying facets, which fine-tune the system (see Figure 1). Li & MacDonnell use the overarching factors, but not the facets, as an unchanging base with an overlying "Social Layer" (giving the NPC an assigned membership in the social order) and an emotion layer, which is similar to Mac Namee's mood layer and does all the changing. In this implementation, three emotions are generated (shame, love, and shock). Through an unspecified interface, the emotion generator changes the emotional state of the character through his interactions with a player. Unfortunately, this system also has a limited set of character responses and an unchanging personality layer, with no long-term effects on the NPC.

| Openness | Extraversion | Neuroticism |
|---|---|---|
| Fantasy | Warmth | Anxiety |
| Aesthetics | Gregariousness | Angry Hostility |
| Feelings | Assertiveness | Depression |
| Actions | Activity | Self-Consciousness |
| Ideas | Excitement Seeking | Impulsiveness |
| Values | Positive Emotions | Vulnerability |
| **Conscientiousness** | **Agreeableness** | |
| Competence | Trust | |
| Order | Straightforwardness | |
| Dutifulness | Altruism | |
| Achievement Striving | Compliance | |
| Self-Discipline | Modesty | |
| Deliberation | Tender-Mindedness | |

**Figure 1. Factors and Facets, Five Factor Model**

No models were found that utilized personality in a truly human way—that is, that included sophisticated base personalities that would develop and change over time based on the NPCs' lived experiences.

## AND THUS THE ENGINE: SUMMARY OF ORIGINAL WORK

In the PPE, NPCs are given adaptive personalities that, through input representing the characters' interactions with a player, change in a realistic fashion over time. This personality development utilizes the Five Factor Model, as did Li & MacDonnell, but digs deeper into the model and uses weighted combinations of the underlying facets to represent various "response types." These response types could be stimuli (the NPC sees that his neighbor has all the latest magical potions, and may feel jealous) or unprovoked action checks (is the NPC gregarious enough to go up to the newly-met player and introduce himself?). Only three such types are used in the PPE (kindness, annoyance, and intimidation), made up of certain facet combinations; for instance, if the kindness response is called, it tests/changes a combination of five facets. As stated in the thesis,

> on the surface, it may seem like channelling the King of the Cosmos in *Katamary Damacy* (Namco, 2004)—'a little WARMTH plus FEELINGS plus a dash of TRUST and a dollop of ALTRUISM minus ANGRY HOSTILITY gives us Kindness'— … in reality these figures are arrived at using correlations between facets and adjectives describing feelings and behaviour from three studies (Costa & McCrae, 1995; Saucier & Ostendorf, 1999; and John et al, 2010), the latter of which is not specific to facets and thus acts more as a tiebreak if the other two disagree.

See Table 1 for an example of these correlations for kindness; note that the strength of the facet's influence on a response type not only determines whether or not the facet

**Table 1. Facet Correlations with Kindness (Only those with Moderate or Above Correlation Shown) (from Georgeson 2011) (all values for PPE)**

| Facet | Saucier & Ostendorf (1999) | Costa & McCrae (1995) | Rating | Weight | Final Weight |
|---|---|---|---|---|---|
| Warmth* | .87 | .41 | Mod-High** | 0.75 | 1.0 |
| Feelings | -- | .43 | Moderate | 0.5 | 0.5 |
| Trust* | -- | .54 | Moderate | 0.5 | 0.5 |
| Altruism* | .70 | .34 | Mod-High | 1.0 | 1.0 |
| Angry Hostility | -.72 | -.41 | Mod-High | -0.75 | -1.0 |

* John et al's (2010) estimated kindness correlation for any facets within Extraversion was .85, and so raised the ratings for those. Their estimated annoyance correlation for Agreeableness was around -.45, and thus moderated those facets within that trait. Also, their estimated kindness correlation for Agreeableness was .74-.80, raising the values for these facets.

**A value of 'Moderate-High' was given to facets falling near the .7 division, or that had wildly disparate ratings from different sources.

is included in the type's calculations, but also provides a weighting system, with low correlation across the studies causing the facet not to be included, moderate correlation allowing inclusion but with a weighting of 0.5, and moderate-high and high correlations providing inclusion and a weight of 1.0. Some facets correlate negatively with a type; for instance, Angry Hostility correlates negatively to kindness (someone with low Angry Hostility is more likely to be kind). These facets are weighted negatively (Angry Hostility is weighted -1.0 for the kindness type).

When a developer calls for information regarding a response type (say, asking whether the shopkeeper NPC is going to be nice to the player entering the store, not really react at all, or perhaps pointedly ignore the player), the PPE finds the facet names associated with that type, retrieves the actual facet values from the NPC's database, and begins calculating a response. We used the PPE to test three different means of calculating responses: one entirely random (as a sort of control; if the NPC's calculated responses made no more sense than a random one, then the engine obviously didn't work), one "Determinism + Probability" (D+P), and one using a modified fuzzy logic system. Each system ultimately returned an integer from 0 to 4, with 0 being the opposite of the response type (the NPC is filled with hostility, in the example of a call for kindness) and 4 being the most extreme example of the response (the NPC is filled with kind feelings, perhaps bordering on love).

The D+P calculation was straightforward: facets and their weights were combined to create a score between 0 and 99. This score was initially compared to a range table to determine a result; e.g., a kindness score of 57 would result in an average response, but a score of 86 would result in a very kind response. Unfortunately, while allowing for changing personalities, this created a very predictable character whose demeanor was easily manipulated (just be kind five times, and the NPC will love you). Thus, a weighted probability system was added so that the NPC was merely more likely to change in certain directions depending on the strength of her facet scores. In this case, a kindness score of 57 would result in a 6% chance of a neutral response and a 94% chance of a slightly kind response, while a score of 86 would give an 8% chance of a slightly kind response, an 88% chance of a very kind response, and a 4% chance that the NPC would immediately offer to help the player. This was better, but it still resulted in discontinuities (because the table of probabilities contained discreet jumps and could result in an NPC being slightly kind one time, but jumping two categories and offering help the next) and limited NPC reactions (i.e., they were still fairly predictable).

In the third system, NPC results were calculated using a modification of fuzzy logic, wherein the weighted facets were combined in such a way as to a) not end up with thousands of possible rules and b) return not a crisp value, but instead a range of possibilities that all could make "sense" to the NPC. Luckily, each facet had the same set of fuzzy linguistic variables (FLVs), based on the ranges in Costa & McCrae's NEO-PI (a test given for the Five Factor model) and the sample NEO Personal Insight Report (Hogrefe 2005): Very Low, Low, Average, High, and Very High. This allowed the facets to be combined into two groups (positive and negative) and thus only two FLVs, and after such grouping and applying the Combs method (which helps to mitigate the problem of combinatorial explosion with additional FLVs by making the increase linear and not exponential; see Buckland 2005), there were a manageable number of rules (10) instead of as many as 9.7 million (for intimidation).

A sort of defuzzification was accomplished by determining the total possible membership confidences (the chances of a result in each of the five ranges) and using this to return the chance of the NPC using one of these values. For example, in one instance the NPC's kind response was figured to have a 0.61 membership in the "High" range (meaning "Kind" in this case, and equating to "Very Kind" in the D+P system) and a 0.29 membership in the "Average" range ("Neutral"). Converting to percentages, the NPC had a 76.25% chance of returning "Kind" (a 3) and a 23.75% chance of returning "Neutral" (a 2). This incorporated a probability system in the same way as the D+P system, but in a much more natural way. It resulted in the smoothest set of results; that is, more reasonable ranges than D+P and smoother personality changes over time, once changes were made part of the system. (Due to space constraints, these results are not reprinted here; see Georgeson 2011, 52-57.)

While interesting, just calling for a response with no change to the underlying facets would fail to change the characters over time. This would be unlike human responses; age studies by Roberts & Mroczek (2008) and Srivastava et al. (2003) show that various personality traits (facets) change over time and thus have a general kind of "elasticity" (see Table 2); larger ranges in the studies equate to greater possible change over time. Thus, for example, a shopkeeper should start to feel more kindly toward his patrons if all were kind to him; his initial response to a stranger entering his shop would be to react more kindly than if he had experienced only neutral or negative interactions with others. In a game environment, these changes are necessarily larger and faster than in real life, but the general idea is the same: creating an "attitudinal memory" that allows the NPC to react appropriately to stimuli over time, even while not remembering specific events.

In the PPE, facets changing a significant amount in the age studies are allowed to change more easily, while those that remain relatively stable in the studies are more difficult to alter. For example, Competence (as part of Conscientiousness) changes a relatively large amount in the studies, and thus is given a high change weight in the engine; Assertiveness (part of Extraversion) changes very little, and thus is given a low weight (Table 2). Note that some items that would normally be given a lower change weight (e.g., Warmth and Gregariousness) were made to be highly changeable, as Roberts & Mroczek included two separate categories for Extraversion, one of which had much higher rates of change than the other, and this highly changeable category could be said to include those two facets. For game purposes, an extra multiplier (controllable by the developer) is added to allow for alterations of the rate of change, so that changes in overall personality can fit into the amount of time spent in the game world and the developer's sense of how quickly the NPCs should change (and each NPC can change at a different rate, if desired).

This prototype system worked well, as far as it went. Three game characters started with vastly different base personalities (see Figure 2) and were approached by a player character who would engage the NPC in a set dialogue routine through a behavior tree. The results of using the D+P and modified fuzzy systems as the basis of NPC responses and changes over time were compared to the responses of a set of the same three characters using only random changes to determine whether the personality-engine based characters seemed to make sense in terms of their responses and, at the very least, could make more sense than a randomly controlled NPC. As would be expected, the NPCs' 30 interactions using a random response were scattered and without a pattern, making little sense in the context of personality development or player-NPC interactions—e.g., from a neutral response to an angry response to a loving response, all in a row and independent of how the player acted.

**Table 2. Amount of Change in Traits over Decades (from Georgeson 2011)**

| Trait | Gender | Range* | |
|---|---|---|---|
| | | Srivastava et al (2003) | Roberts & Mroczek (2008)** |
| Openness | M | 77-73 | 0.2-0.8 |
| | W | 74-71 | 0.2-0.8 |
| Conscientiousness | M | 58-69 | 0-1 |
| | W | 62-71 | 0-1 |
| Extraversion | M | 52-53 | 0.1- -0.1*** |
| | W | 57-56 | 0.1- -0.1*** |
| Agreeableness | M | 64-68 | 0.0-0.7 |
| | W | 66-74 | 0.0-0.7 |
| Neuroticism | M | 46-45 | 0.2-0.9 |
| | W | 58-49 | 0.2-0.9 |

\* Range in Srivastava et al is based on percentage of maximum score possible over four decades (ages 21 - 60); range in Roberts & Mroczek is in numbers of standard deviations over roughly six decades (~15 - ~75).
\*\* Roberts & Mroczek did not divide their study by gender; thus the same score is given to both.
\*\*\* Extraversion in Roberts & Mroczek is divided into two groups, Social Vitality and Social Dominance. The scores given are for Social Vitality; Social Dominance, which could be said to include Gregariousness and Warmth, had a much greater variance (0.2 - 1.1).

**Figure 2. Initial Facet Values for Three NPCs (SS = Shaman Shopkeep, TT = Tilla Transit, AG = Alan Guardsman) (from Georgeson 2011)**

| Openness to Experience | SS | TT | AG |
|---|---|---|---|
| Fantasy | 20 | 75 | 30 |
| Aesthetics | 20 | 70 | 20 |
| Feelings | 55 | 75 | 30 |
| Actions | 30 | 80 | 20 |
| Ideas | 20 | 70 | 20 |
| Values | 20 | 60 | 10 |

| Conscientiousness | SS | TT | AG |
|---|---|---|---|
| Competence | 70 | 50 | 80 |
| Order | 80 | 30 | 80 |
| Dutifulness | 50 | 30 | 90 |
| Achievement Striving | 60 | 35 | 75 |
| Self-Discipline | 80 | 25 | 85 |
| Deliberation | 50 | 25 | 50 |

| Extraversion | SS | TT | AG |
|---|---|---|---|
| Warmth | 55 | 85 | 40 |
| Gregariousness | 45 | 70 | 30 |
| Assertiveness | 50 | 40 | 30 |
| Activity | 50 | 75 | 50 |
| Excitement Seeking | 30 | 70 | 30 |
| Positive Emotions | 50 | 80 | 50 |

| Agreeableness | SS | TT | AG |
|---|---|---|---|
| Trust | 40 | 60 | 40 |
| Straightforwardness | 70 | 70 | 70 |
| Altruism | 40 | 70 | 30 |
| Compliance | 50 | 50 | 65 |
| Modesty | 50 | 50 | 70 |
| Tender-Mindedness | 50 | 70 | 25 |

| Neuroticism | SS | TT | AG |
|---|---|---|---|
| Anxiety | 45 | 35 | 20 |
| Angry Hostility | 50 | 30 | 65 |
| Depression | 50 | 50 | 20 |
| Self-Consciousness | 50 | 25 | 45 |
| Impulsiveness | 20 | 70 | 20 |
| Vulnerability | 50 | 40 | 20 |

In the D+P and fuzzy systems, the NPCs' responses (and, more importantly, their possible ranges of response) to player interactions develop in a more rational way; there is more and more chance of an otherwise kind character eventually throwing a repeatedly belligerent player out of the shop, as she grows more and more unhappy with the interaction over time. Conversely, NPCs who start with a less-kind attitude toward others (e.g., a guard such as Alan Guardsman) reach this stage much more quickly, and have a more difficult time developing very kind or loving feelings toward others. As stated before, while both systems worked, the fuzzy system provided somewhat more realistic interactions and change in personality (again, see Georgeson 2011 for details). There were a few issues with speed of change (which was adjusted over the tests) and with some of the intimidation responses (resulting from trying to do too much with that facet alone), but the system seemed ready for further development.

**POST-2011 PERSONALITY ENGINES (BESIDES ExAI)**

There have been several developments since the creation of the PPE that directly involve NPCs and personality. Bura et al. gave a presentation at the Game Developer's AI Summit in 2012 which talked, in part, about using the underlying facets of the Big Five in order to give NPCs personalities. Bura's idea is to give +x/-x (between 2 and -2) values to combinations of facets to create needs and behaviors (collectively called traits), much like ExAI uses combinations of facets to create the response/stimulus types. However, a big difference is that characters do not have sets of values for their own facets; instead, the traits are each packaged as a set of values and are given (tagged) to a character. For example, the "Shyness" trait can be created through a set of adjustments to various facets. This trait is then tagged to a given character to make her shy. Bura compares different traits and creates a scalar product to help determine the NPC's course of action (if the product is high enough, then the behavior will occur). Bura explicitly states that he is not going for realism here, but for "the illusion of inner life" and a straightforward way of determining character actions. There is no indication that these values change over time, or that the NPC can have differing and adjustable values in regard to different players, other characters, etc. This GDC talk influenced the development of Pixel Crushers' Love/Hate (2015).

Love/Hate is similar to ExAI in that it creates personalities for NPCs, but it focuses on changes to relationships and, like Li & MacDonnell (2008), emotions, in this case using the PAD (Pleasure-Arousal-Dominance) model for NPCs' emotional states, and using a number of "templates" (including the factors and the entire facet set of the Five Factor model) for static underlying personalities. (PAD was first proposed by Mehrabian & Russell (1974) and scores emotional states on a three-dimensional scale of Pleasure-Displeasure, Arousal-Nonarousal, and Dominance-Submissiveness.) The emotional values range from -100 to 100 and are the primary drivers in character changes. Personality-wise, +/- values (in this case ranging from -100 to 100) are used to create an overall personality. These personalities are used in Love/Hate as faction templates; that is, they can be used as starting points for a social circle of NPCs (e.g., all elves from a certain village start with the same template). Unlike in Bura's model, in Love/Hate each character has his or her own set of facet values.

The biggest differences between Love/Hate and ExAI are in how emotions are handled and in how personalities change over time. In Love/Hate, emotions are ever-changing variables that don't affect the underlying personality; in ExAI, they are dealt with as part of the facet system, potentially (if strong enough and repeated enough) altering the NPC's personality and feelings toward the specific player and, to a lesser extent, toward everyone she meets. And in Love/Hate, the personality facets are unchanging—what the NPC is born with is how she remains—while one of the main points of ExAI is the changeability of personality over time.

A different kind of engine altogether is Versu (Evans & Short 2014; Short 2013), focusing on social interaction between characters. This is much more about the ways in which social interaction can take place than it is about utilizing personality theory. The NPCs' speech itself (as heard and overheard by others) causes the opportunities for change in characters' interactions with and attitudes toward one another. However, in a sense each NPC does have a personality, represented by desires; as Evans states, instead of using a finite set of personality traits, they wanted "a more expressive system, in which there were an *infinite* number of personalities—as many personalities as there are sentences in a language" [italics in original]. Rather than generalize from a personality trait, as one would do with ExAI, a character has specific attributes, such as being sexist or hating to be alone.

Additionally, characters evaluate those around them through role-evaluations; that is, through how well a character is playing a social role. In Versu's Jane Austen episodes, for example, such evaluations include how well-bred someone is, how properly they are behaving, how attractive they are, and so on. These evaluations (and character rela-tionships) change over the course of the simulation depending on character interactions. So while their "personalities" are not changing, their thoughts and feelings toward other characters are.

Versu speaks directly to belief systems; that is, the NPCs know social norms (as represented in the game) and will, in general, follow them. The ExAI engine does not deal with belief systems, except insofar as the underlying personality would affect the beliefs of the character—but for now, the developer would have to create these links herself.

Comme il Faut (CiF; McCoy et al. 2011), used in Prom Week, is another social engine, similar to Versu in using characters' social knowledge and rules to determine interactions between NPCs. It models traits that describe perceptions (e.g., "attractive," "weakling"), as opposed to internal personality (which can only be perceived by others at a remove); CiF's traits are thus more like ExAI's stimulus/response types. Also, CiF's engine is used to keep the NPCs functioning within the social parameters of the world, whereas ExAI focuses on individuality.

Interestingly, all three of these latter engines could be integrated in some way with the ExAI system, creating more dynamic overall systems that could make NPCs even more human.

## PUTTING THE EXTREME IN THE AI: POST-PPE DEVELOPMENTS

During the last four years ExAI has gone through extensive development. As one would expect, many changes were made in bringing ExAI from a master's project to a marketable, robust personality engine. The major changes are discussed below.

### Many, many stimulus/response types added

To realistically simulate human personalities, facets needed to be combined into far more than the three stimulus/response types tested in the thesis. Using the same research as that for the original three types, 39 types were eventually defined that seemed, in testing, to be the most useful (Figure 3). Facet weighting was also determined using the same research as used for the original three types.

While these types cover many situations/interactions, 39 is not a magic number; indeed, some seem to cover different aspects of the same thing (happiness/sadness), while there are certainly other responses that could be created (one could come up with a virtually unlimited number). Work is ongoing to refine this list.

| | | |
|---|---|---|
| sadness | deceitfulness | assertiveness |
| happiness | condescension | humour |
| anger | quarrelsome | intellectual |
| sternness | helpfulness | demanding |
| jealousy | selfishness | productive |
| anxiety | affectionate | ambitious |
| impulsive | dependability | orderly |
| stubborn | efficiency | kind |
| guilt | moodiness | annoying |
| standoffish | wittiness | intimidating |
| reluctance | excitability | gregarious |
| conformity | imaginative | |
| distrustful | talkative | |

**Figure 3. Stimulus/Response Types in ExAI**

As a sort of "special" type for *They Vote!,* two more response types were created attempting to simulate voters' conservative or liberal leanings and their tendency to either vote or not vote in elections. While certain facets do indeed help to determine such ratings (especially Values), this would likely work better integrating various other systems as well (especially memory and morality, discussed in The Future of ExAI, below).

**NPCs have different attitudes/reactions to different entities**

In the PPE, a character's personality changes were generalized: A shopkeeper's attitude changed toward all customers an equal amount. If one customer was mean, the NPC would treat every customer as though he or she had been mean. (This did not matter in the original testing because the NPCs only faced one person repeatedly.) In ExAI, a character's personality changes are individualized; that is, her attitude toward the entity causing the change is altered more than her attitude in general. So she may begin as generally neutral, with no real like or dislike for her customers. If one player comes in often and is always nice to her, her attitude toward that player will change significantly, and her overall feelings toward new customers will skew slightly as well (although not nearly as much). Conversely, if another player starts coming in and is always nasty to her, her attitude toward that second player will change significantly in the opposite direction, and her overall feelings toward strangers will adjust back toward her original feelings (or get worse; she may feel slightly betrayed by her own feelings)—but her attitude toward the first player will continue to be significantly positive.

Or say an opposing computer-controlled manager in a football management game is generally distrustful and selfish. It might be possible to, over time, change this distrust enough so that the manager trusts you in his dealings with you, and possibly could be made a little less selfish. This would not necessarily change him in his dealings with others, however, and he could have quite different opinions of each of them.

**Personality Storage**

Originally, the facets for each of the NPCs were stored in a SQLite database using several interrelated tables; while this worked for the original project, two issues confounded attempts to use this method in a game: speed of access and lack of support across all platforms. As discussed below, facets are accessed in many combinations for each method call, and it can be necessary to read and write to these facets for many characters at a time. For instance, in the *They Vote!* demo (see the next section), 100 NPCs were polled at once whenever a politician made a speech or acted in some other way that might influence the voters. Accessing everything using the database, this could take over 3 minutes to complete. And even in *SteamSaga*, where conversations are generally one-on-one, given the demands of graphic rendering, physics calculations, audio, etc. a complex facet check could occasionally create a slight "hiccup" in movement—which in an action game is untenable.

In addition, using a database, even SQLLite, creates support problems across different platforms. For instance, Apple computers wouldn't recognize the database at all.

Our solution is to store character personalities in encrypted XML files, and to load entire personalities into memory during the initial game load (or level loads, etc). Here the personalities are manipulated and accessed; they are saved back out to the XML files only as requested, and reloaded only as requested. And as these are not large amounts of text, they require very little overhead.

This solution worked exceedingly well. In exchange for the robustness and ease of manipulation of a relational database, speeds were increased at least 200-fold, as the *They Vote!* polling now took less than a second, and there were no longer any hiccups in *SteamSaga*. And now anything that could read and write to an XML file could use ExAI—making it cross-platform.

**Fine-tuned the facet adjustments**

With additional response types came additional testing. Lots of testing. This lead to a better understanding of the "right" adjustments for the various facets over time. There is also, however, a built-in way for developers to alter the facet change rate, both in the way an NPC reacts to individual entities and in his reactions to everything. Changing these rates for one NPC has no effect on other NPCs.

**TWO GAME DEMOS—*THEY VOTE!* AND *STEAMSAGA***

Quantum Tiger Games has used ExAI in two game demos. *They Vote!* (2014) is a basic voting simulation in which one hundred voters (represented by differently colored circles on an island) are wooed by two political candidates, one conservative and one liberal. Over time, as the player chooses actions and responses for the politicians, the voters may or may not change their minds about who they believe to be the best candidate. *They Vote!* deals with the effects of personality on individual decisions, and it also might have some bearing on the behavior of crowds (although it wasn't set up to measure this).

The voters begin with a wide range of personalities, including ultra-conservative, leaning conservative, neutral, leaning liberal, and ultra-liberal. To simulate the politicians having been around for a while (and thus the voters already having some knowledge and opinion of them), a choice is given to the player having to do with the politicians' personalities. After the effects of these personalities on the voting public having been calculated, a poll is taken as to who would vote for whom. After this, there are several situations from which the player can choose. Given the same stimuli, all the voters' personalities (and thus voting behaviors) are adjusted, given their base personalities and how they've grown to feel about the politicians.

Voters are tested across three variables: how they feel politically (from conservative to liberal), which candidate they like best (feel most friendly toward), and which candidate they trust the most. They are also tested for whether or not they are likely to vote at all (given certain personality aspects having to do with apathy, etc.). They are then polled again and the results shown.

For example, in the beginning a poll is run based only on the voters' political leanings, as though no candidate at all were running. This will generally show the more extreme blocks of voters and most of those who lean one way or the other adhering to their belief systems. The undecided block has few who declare for either party, as does the group who needs great impetus to vote at all.

In the next round of polling, candidate backgrounds are added. The player chooses from among several options; e.g., "Jackson very likeable, efficient, not very dependable; Kingston unlikeable, highly efficient, trustworthy." These candidate traits are digested by the voters; or, rather, each voter "interacts" with each candidate trait and has a reaction to it, based on personality. If a candidate seems trustworthy, the voters will likely feel a positive sense of trust in that candidate (modified by their political leanings and those of the candidate). And so on.

Each run-through of this simulation provides a slightly different result, even if the same options are chosen by the player; this is to be expected, as the people (NPCs) involved may not have exactly the same strength of feelings in each alternate universe. Those who feel strongly will be the same, but because of the fuzzy logic built into the system, those who are on the fence may teeter one way or the other. This is very much on purpose; real people are not entirely predictable, and thus the NPCs shouldn't be, either. For example, in three sample runs, the conservative candidate is given the most advantage: He is by far the most likeable, while his opponent is unlikeable and somewhat untrustworthy; only his promises and opinions are chosen (and in these runs only the most believable of these); and no political mudslinging from either side takes place. In general, this has the desired effect (from his point of view): in each run, he wins by around 30 points, no matter that the liberal has varying leads of one to eight points from the initial round (Table 3). However, the exact numbers are different in each run; both he and the liberal candidate keep their core constituencies, by and large, but the undecideds and those least likely to vote have slightly different opinions of the candidates in each alternate world.

*SteamSaga* (2014) is a demo of an RPG with four nameless characters—a Fighter, a Healer, a Thief, and a Bard—on a mysterious journey. They begin without knowing their own names, where they're going, or why. They're also unsure of their relationships to one another. Using ExAI, they develop (or begin to develop, as the demo is only about ten minutes long) feelings about and attitudes toward one another. For example, in the opening conversation, the player's Fighter character is asked by the Healer whether he recognizes any of the objects nearby. The player can answer honestly, elusively, or not really answer at all. If the player answers elusively, the Healer may pick up on this and begin to feel anger toward the Fighter (that is, ExAI runs a method that increases her anger toward the Fighter, which affects all the underlying facets of which anger consists). She will try to get the Fighter to answer again, and if the player continues to be elusive, she will get angrier—which doesn't have an immediate effect (she's not explosively angry), but in future conversations she may treat the Fighter less kindly, and with less respect, depending on the results of personality checks. Note that her anger toward the Fighter increases her general feelings of anger (but not as much), and that repeated angry encounters with group members may raise her tendency to be angrier in general.

Battles also offer a chance for the engine to be used: While the player is nominally the leader and controls what each character does during battle turns, each NPC has preferred (and hated) things to do/not do during battle. For instance, if the player-controlled Fighter sends the Healer into a physical battle (even after she's stated that she prefers to stay back and cast spells), she'll begin to dislike the Fighter and may eventually decide that the player's leadership abilities are suspect. She can decide not to do what the player tells her to do, doing her own thing in battle situations. This also affects situations outside of battle—her loss of respect toward the player is reflected in conversational choices thereafter, unless the player somehow wins it back.

## THE FUTURE OF ExAI

In general ExAI is a robust solution as far as it goes, but as with any software it could be fine-tuned, or additions could be made to make it even more realistic. After all, humans don't consist of personality alone. The following are some additions/modifications that could be made:

*Adding an emotion layer for strictly transient emotions,* similar to Love/Hate (2015), Mac Namee's mood engine (2004), Li & MacDonnell's Emotion layer (2008), and others. This could add more depth, although thus far transient emotions can be mimicked successfully using the engine as-is (e.g., reversing whatever facet changes were involved in the emotion when the stimulus is gone).

*Integrating a memory system.* This is, in fact, in progress (Georgeson, forthcoming). Currently, the ExAI engine "remembers" how an actor has made the NPC feel or otherwise affected the personality of the NPC through the actor-specific adjustments to facets, but it does not remember specifics (e.g., Clyde knows that the player has been a source of annoyance to him in the past, but couldn't tell you exactly what happened to cause this). A memory system would add these details and associate them with any actors or situations involved, in addition to being affected by the character's personality (as, for instance, working and short-term memory can be affected by depression, anxiety, and other characteristics [see, e.g., Kizilbash, Vanderploeg, & Curtiss 2002]).

Table 3. *They Vote!* Results, Generally Favorable to the Conservative Candidate

| First Run | | | | Second Run | | | | Third Run | | | |
| --- | --- | --- | --- | --- | --- | --- | --- | --- | --- | --- | --- |
| Round | Vote Con | Vote Lib | Und | Round | Vote Con | Vote Lib | Und | Round | Vote Con | Vote Lib | Und |
| Initial | 26 | 27 | 47 | Initial | 23 | 31 | 46 | Initial | 22 | 29 | 49 |
| Personality Chosen | 32 | 18 | 50 | Personality Chosen | 36 | 16 | 48 | Personality Chosen | 32 | 26 | 42 |
| 1 | 33 | 18 | 49 | 1 | 37 | 24 | 39 | 1 | 37 | 19 | 44 |
| 2 | 40 | 15 | 45 | 2 | 43 | 17 | 40 | 2 | 42 | 12 | 46 |
| 3 | 46 | 14 | 40 | 3 | 38 | 14 | 48 | 3 | 46 | 17 | 37 |
| 4 | 48 | 16 | 36 | 4 | 49 | 13 | 38 | 4 | 44 | 15 | 41 |
| 5 | 48 | 17 | 35 | 5 | 49 | 15 | 36 | 5 | 44 | 17 | 39 |
| Final Tally | 48 | 17 | 35 | Final Tally | 49 | 15 | 36 | Final Tally | 44 | 16 | 40 |

*Integrating a natural language system.* A bit of a holy grail, but such a system would make everything even more realistic. As stated in the original, "having the NPC respond to and understand natural language would be a great boon to the realism of any game" (Georgeson 2011), and we're much closer to seeing viable natural language systems now (e.g., Siri, Cortana, Google Now), but we haven't reached a point where the AI can fully understand what's being said to it. Such a deeper understanding would be key to tying the personality engine and natural language system together.

*Integrating a morality system.* Perhaps this should be "a morality and ethics system," which would need to be both moral and ethical. Such a system used in a game could really flesh out the NPCs, but has the same potential problems as the personality engine insofar as the characters could act in very unpredictable ways over time. However, as with the personality engine, the developer would not have to worry about the NPCs' taking over the game a la the 1973 film *Westworld*, unless said dev really wanted it to be this way.

## CONCLUSION

The ExAI personality system, as fully developed, has met expectations that it could provide a human-like personality, changing over the course of a game character's existence—helping these NPCs really live in their game worlds rather than be part of the scenery. The engine worked well in both an RPG and a simulation, and would likely work just as well in any game with non-player agents: sports or other management simulations, strategy games like the Civilization series, and so forth. Further tests will be done in terms of player reactions to ExAI-driven NPCs.

ExAI is but a beginning, however, for it is part of a set of larger projects to make game characters and other agents as realistic as possible, including memory, morality, natural language, and more.

## ACKNOWLEDGMENTS
Thanks to Dr. Christopher Child, who was the thesis advisor for the original PPE thesis, and is thus listed as co-author of this paper.